# A Sagittal Planar Ankle-Foot Prosthesis with Powered Plantarflexion and Socket Alignment

Mark A. Price, *Member IEEE* and Frank C. Sup IV, *Member, IEEE*

*Abstract*—Powered ankle-foot prostheses can often reduce the energy cost of walking by assisting with push-off. However, focus on providing mechanical work may lead to ignoring or exacerbating common issues with chronic pain, irritation, pressure ulcer development, and eventual osteoarthritis in persons with amputation. This paper presents the design and validation of a novel transtibial prosthesis informed by predictive biomechanical simulations of gait which minimize a combination of user effort and interaction loading from the prosthesis socket. From these findings, the device was designed with a non-biomimetic anterior-posterior translation degree of freedom with a 10 cm range of motion which is primarily position-controlled to change the alignment of the prosthetic foot with the residual limb. The system is both mobile and tethered, with the batteries, actuators, and majority of electronics located in a small backpack. Mechanical loads are transmitted through cables to the prosthesis, minimizing the distal mass carriage required. We measured torque and force sensing accuracy, open loop actuator performance, closed loop torque and position control bandwidth, and torque and position tracking error during walking. The system is capable of producing up to 160 N-m of plantarflexion torque and 394 N of AP translation force with a closed loop control bandwidth of about 7 Hz in both degrees of freedom. Torque tracking during walking was accurate within about 10 N-m but position tracking was substantially affected by phase lag, possibly due to cable slack in the bidirectional mechanism. The prototype was capable of replicating our simulated prosthesis dynamics during gait and offers useful insights into the advantages and the practical considerations of using predictive biomechanical simulation as a design tool for wearable robots.

## I. Introduction

PEOPLE living with major lower limb amputation are much more likely to suffer from chronic pain and secondary impairments than able bodied individuals [1], [2]. In addition to causing discomfort and distress, post-amputation chronic pain may contribute to higher energy costs and reduced mobility [3], may be a precursor to comorbidities such as pressure ulcers and osteoarthritis [1], [4], [5], and can lead to prosthesis abandonment and reduced physical activity overall [1].

A major cause of this pain and discomfort is excessive pressure exerted by the socket on parts of the residual limb [6]. These pressure hotspots are largely the result of transmitting large ground reaction forces as moments through the prosthesis socket and into the residual limb [6], [7], [8], [9]. Across a typical anatomical joint, moment loads are transmitted to neighboring segments through the muscles and connective tissue, which act as tension members and keep the skeletal structure largely in compression along the long axis of each segment. Across a socket-residuum connection, however, no tension members are present, and so moments induce reaction loads against non-loadbearing tissue [10].

While developments in socket design have shown promise in improving the distribution of loads on the residual limb and forming stiffer connections between the anatomical and artificial limbs [11], [12], [13], they cannot directly change the magnitudes and load paths of the ground reaction forces through the socket. This can, however, be accomplished by controlling the behavior of the artificial limb when it is in contact with the ground. Torsional and compressive loads have been shown to be attenuated by the addition of passive compliance in the corresponding axes at the artificial ankle joint [14], [15]. Reducing pressures due to sagittal plane bending moments in the same way may not be a viable solution, as it would dissipate the ankle push-off work that provides most of the positive joint power during normal walking [16], [17].

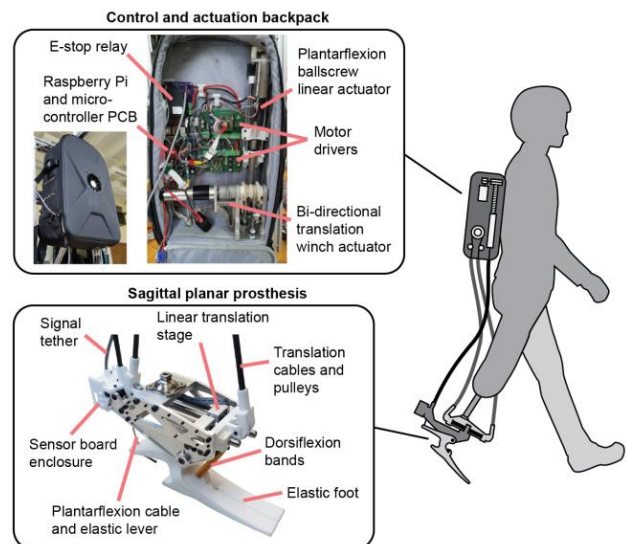

Figure 1. System overview: Actuators and computation are housed in a portable system worn on the torso and tethered to the prosthesis. Powered plantarflexion and bi-directional anterior-posterior translation are enabled through Bowden cable transmission.

This work was supported by the National Robotics Initiative with a grant from the National Science Foundation (IIS-1526986)

M. A. Price and F. C. Sup IV are with the University of Massachusetts, Amherst, MA 01002 USA (e-mail: {mprice, sup}@umass.edu).

There is evidence that active control of the anterior-posterior (AP) alignment of the foot with respect to the tibia can reduce socket flexion moment while preserving the push-off function of the ankle. A powered ankle-foot with coupled dorsi-/plantarflexion and AP translation demonstrated reduced pressures inside the socket compared with the passive daily-use prosthesis for a single pilot participant with unilateral transtibial amputation [18]. Additionally, biomechanical simulations of a human walking with a robotic ankle-foot with powered plantarflexion and AP translation minimized combined objectives of total muscle effort and peak socket loading [19], [20]. These simulations predicted that both peak socket moment and cumulative effort spent by the muscles can be reduced with active AP translation as compared to a powered revolute ankle joint.

Despite promising experimental and simulated outcomes, additional control of AP translation at the ankle joint is non-anatomical and may come with practical or preferential tradeoffs in human users that are not apparent in simulations or basic mechanical analyses. While initial experiments with a prototype prosthesis indicate improvement relative to an unpowered prosthesis, there is no existing experimental tool for assessing the socket load reduction, user preference, and overall biomechanical changes in real human gait caused by varying offload translation magnitudes in an otherwise normal revolute artificial ankle joint.

To fill this gap, we designed and prototyped an artificial ankle-foot with independently controlled plantarflexion and AP translation degrees of freedom (DoF). Our objective was to create a test platform for experimental research of human gait with independent control over ankle plantarflexion and AP translation of the foot capable of replicating our simulated behaviors [20]. To maximize the configurability of the test platform and allow for both overground and treadmill testing, we designed the prosthesis and actuation system as fully modular wearable subassemblies.

In this paper, we present the mechatronic design and control of the prosthesis test platform (Section II). We evaluate the mechatronic performance of the system with a series of bench tests and a pilot walking test with one able-bodied participant wearing prosthesis adapter boots (Section III). We then discuss the main findings of the evaluation tests, the limitations of the design and suggested improvements, and the potential follow-up studies and applications of the device (Section IV). We conclude by summarizing our main contributions and opportunities for future work (Section V).

## II. MECHATRONIC DESIGN

In this section, we provide a comprehensive overview of the prosthesis test platform. We describe the mechanical design of the artificial limb, the design of the sensing, actuation and computing system, and the design of the control modes.

### A. System Overview

An overview of the complete prosthesis system is illustrated in Fig. 1. To minimize the encumbrance of the artificial limb while enabling overground locomotion, we designed the prosthesis system into two modular subassemblies: 1) A prosthetic ankle foot with independent AP translation and dorsi-/plantarflexion DoFs, and 2) A back-mounted enclosure containing the actuators, the embedded computing system, and the majority of the electronic hardware. Mechanical power from the actuators is transmitted to the prosthesis via Bowden cables. The prosthesis is instrumented with a 3-axis accelerometer and position sensors for each DoF. Force and torque are not sensed directly on the prosthesis, but are computed from measurements of the stiffness-characterized actuator transmission deflection. Sensor data and control signals communicate along a tether cable between the prosthesis and the control backpack.

### B. Prosthesis Design

The ankle-foot prosthesis consists of three subassemblies with two modular DoFs: a prismatic joint (A-P displacement) and a revolute joint (ankle dorsi-/plantarflexion). The prismatic joint is constructed as a stage riding on linear bearings across a pair of shafts. The shafts are mounted into a rigid truss which houses the pyramid connector attachment point for a standard prosthesis pylon. The revolute joint is constructed as a set of needle bearings and rotary shaft mounted below the linear stage. The rotary shaft is press-fit into a custom 3D printed prosthetic foot. The prosthetic foot was created using selective laser sintering with nylon powder (Formiga P110, EOS). The resulting solid material is highly elastic and robust, allowing the foot to function similarly to an elastic carbon fiber foot. The linear and rotational subassemblies can be decoupled and the prosthesis pylon connection can be relocated to the rotational subassembly, allowing it to function as a 1-DoF revolute ankle.

Both prosthesis DoFs are actuated via synthetic cables (West Marine, V-12 Vectran Single Braid) through coiled-steel conduits (Lexco, 415310-00) attached to remote actuators. Two cables are routed through pulleys and attached to the front and back of the linear stage, capable of applying bi-directional force through agonist-antagonist action. One additional cable attaches to a torque-amplifying lever built into the prosthetic foot and exerts plantarflexion torque by applying cable tension upward. Dorsiflexion torque is generated passively by elastic bands connecting the front of the device frame and the top of the prosthetic foot.

The prosthesis mass and ranges of motion are reported in Table 1. Note that the translation range of motion is constrained primarily by the allowable length of the "tail" section of the prosthesis frame (Fig 2), which routes the plantarflexion cable behind the limb. The dorsi-/plantarflexion range of motion is primarily constrained by the allowable height of the device.

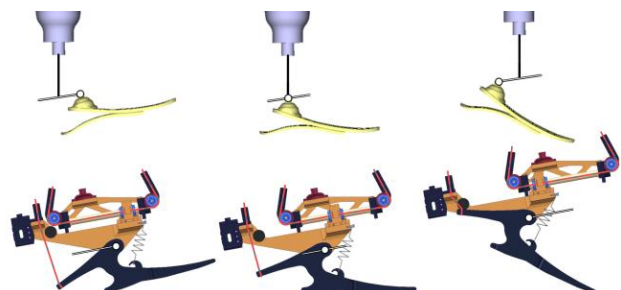

Figure 2. Illustration of mechanism motion. Top row: Device range of motion limits in OpenSim simulations used to generate prosthesis design parameters. Bottom row: Corresponding range of motion in prototype design.

### C. Actuation and Control Pack Design

Given the actuation requirements and mechanical complexity required by the prosthesis itself, the actuators and control electronics are located in a separate enclosure. Instead of locating these components fully offboard, as with many of the emulator devices which seek to minimize mass worn by the user, they are stored in a backpack. This design choice shifts much of the mass of the full assembly near to the center of mass of the body, mitigating the effects of the added weight while remaining untethered to an offboard platform. While this limits the size of the actuators that are practical for use, and therefore the power available to the prosthesis, it increases the flexibility of the platform for more varied testing conditions (e.g., navigating overground, an instrumented obstacle course, etc.).

The actuation and control module contains a custom central embedded system, an emergency stop relay, and two actuators connected to Bowden cable outputs. The embedded system consists of two layers: 1) A microcontroller (dsPIC33F, Microchip) which performs low-level operations requiring precise timing and high speed (e.g. analog to digital conversion, digital signal processing, polling sensors at >1 kHz), and 2) A Raspberry Pi 3.0 which handles high-level control instructions and communicates with a host PC via wireless internet to stream data for inspection and logging and receive operational commands in real time via MATLAB Simulink. This system is open source; microcontroller firmware, Simulink models and code, and circuit board design files can be found in a public repository included in the supplemental information.

Actuator commands are output from the microcontroller and handled by separate motor drivers (Maxon, ESCON 50/5), which drive a pair of 200W brushless DC motors (EC-4pole 30). The motors are connected to two custom actuator mechanisms corresponding with the two ankle-foot DoF: 1) A linear actuator which provides unidirectional cable tension to power ankle plantarflexion, and 2) a bi-directional winch to power AP translation. The linear actuator is constructed with a miniature ballscrew (MRT 8X2.5, Nook), chosen to efficiently achieve the gear ratio required to transmit high loads required for ankle plantarflexion. Plantarflexion torque is generated by contracting the linear actuator, which creates tension on the Bowden cable, whereas dorsiflexion is achieved by creating slack by extending the linear actuator and allowing the elastic bands on the ankle-foot to pull the line taut. The bi-directional winch consists of two Bowden cables wrapped in opposite directions around a drum, which is coupled to the motor with a 23:1 planetary gearhead (GP 32 HP, Maxon). Cable tension is maintained with spring-loaded idler pulleys to minimize backlash. This configuration provides a smaller gear reduction than the linear actuator due to the relatively small AP loading on the ankle compared to the plantarflexion moment, allowing for high-bandwidth bi-directional positioning of the foot. Both actuators are designed to meet the torque, velocity, and power requirements indicated by biomechanical simulations (Fig 4E and 4J)

TABLE 1. PROSTHESIS DESIGN PARAMETERS

| Parameter | Value |
|---|---|
| Mass (kg) | 2.29 |
| Build height (cm) | 19.5 |
| Flexion range (deg) | 15 dorsi. to 38 plantar |
| Translation range (cm) | -5.0 to +5.0 |

The actuation and control pack has an input voltage range of 6-48V and maximum current draw of ~30A (15A per motor). This supply voltage may be provided by a set of lithium-polymer batteries or an offboard power supply. All components are mounted to a panel attached to the back wall of the backpack, with load cables and a signal tether passing through the bottom surface. A handheld emergency stop is provided to allow the user to immediately cut power to the motors while leaving the low-power electronics running. A cooling fan and ventilation slits are included to prevent the electronics and motors from overheating. The backpack measures 44.5 x 30.5 x 15.2 cm and has a mass of 6.29 kg including the cable assembly and batteries.

### D. Sensing

A miniature magnetic rotary encoder (RM08, RLS) with 1024 count (0.35°) resolution is mounted on the ankle flexion axis shaft to measure the ankle dosri-/plantarflexion angle (RM08, RLS). A linear magnetic scale with 20 µm resolution is mounted on the truss frame, with corresponding incremental magnetic encoder (RLC2IC, RLS) mounted on the moving linear stage to measure the linear displacement. A 3-axis accelerometer (ADXL335, Analog Devices) is mounted to a custom circuit board at the rear of the prosthesis for the detection of heel-strike during walking. A PIC microcontroller (dsPIC33FJ64MC202, Microchip) located on the custom board processes the sensor data and routes it to the wearable off-board system for logging and controller feedback via signal cable tether. Each motor is additionally equipped with a 512 count rotary encoder (Encoder MR, Maxon). Measuring motor position and output position allows the measurement of stretch or deformation of the actuator transmission.

In order to reduce the number of components and the mass of the device, translation force and plantarflexion torque are estimated by modeling the elasticity of the system and measuring the deflection of the end effector relative to the motor position as opposed to implementing traditional strain gauge load cells. This elasticity is designed into the flexion transmission chain as a leaf spring lever at the rear of the prosthetic foot. Force sensing in the translation axis is designed as an approximate measure, rather than as a feedback signal to a force controller, due to the indication of prior simulation work results that translation is primarily used to position the residual limb advantageously rather than generate substantial positive work ([20], also refer to Fig 4H). Therefore, no additional series elastic component has been introduced to the translation kinematic chain. The Vectran cable itself does stretch under load, however, allowing for some estimation of cable tension.

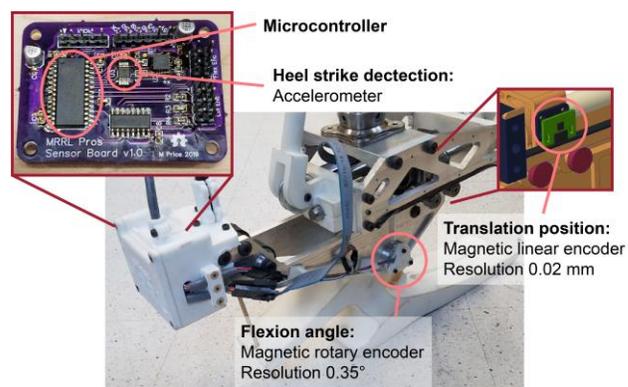

Figure 3. Sensing capabilities and sensor placement on the prosthesis.

### E. Control

The prosthesis controllers are designed in a hierarchical structure. At the top level, walking events are detected using the onboard sensors and used to identify the phases of progression through the gait cycle as they happen. This information is used to determine the virtual stiffness and setpoint of a stiffness controller corresponding with the gait phase. The stiffness control commands an output force or torque determined by the mechanism deflection from the assigned setpoint in imitation of a passive spring. This desired force or torque command is sent through a force/torque control loop. Alternately, the commanded setpoint is used as the reference target for a position control loop.

This control scheme is designed to replicate prosthesis dynamics which minimize human effort and socket-residuum interaction loads in an updated set of forward simulations of gait we previously published [20], [21]. Relative to this work, we updated the prosthesis model from a generic 2-DoF actuated joint to a model of the actual prototype mechanism in OpenSim 3.3 [22], including its actual weight and actuator limits, and generated optimal strides using custom scripts in Matlab to implement a direct collocation trajectory optimization using the IPOPT solver [23]. Other details regarding the model definition, optimization method, and objective functions are identical to [20]. In brief, this simulation generates periodic walking strides that minimize a weighted sum of muscle effort, represented as the time integral of cubed muscle activations summed across all muscles, and socket loading, represented as a smooth approximation of peak compressive force and bending moment on the residual tibia.

Furthermore, we programmed a revolute-only control mode which replicates the active plantarflexion control of other cable-actuated prostheses [24], [25], [26] to demonstrate the flexibility of the design and validity of its performance relative to the field. The prosthesis dynamics resulting from gait simulations are illustrated in Fig. 4.

*1) Walking Control*

The top level walking controller is a finite-state machine which identifies standing and walking states based on detection of gait events. During walking, the controller further distinguishes between stance and swing phases. The prosthesis applies the plantarflexion torque and translation profiles during the stance phase. During swing, both DoFs transition to position control, which lifts the toe and centers the forward translation to avoid collision with the ground. Just before the next anticipated heel strike, the prosthesis resets to its starting configuration for the beginning of the next stance phase.

*a)  Gait event detection*

The onboard sensors allow for gait events to be detected in two ways: 1) rapid changes in prosthesis acceleration or jerk (time derivative of acceleration), and 2) rising and falling edges of the ankle torque signal. The controller identifies heel-strike when jerk is recorded above a threshold, heuristically tuned to the user. It identifies toe-off by a falling edge of plantarflexion torque past a similarly tuned threshold, which occurs when the wearer breaks contact with the ground and the actuator suddenly encounters no resistance. These events mark the start and end of the stance phase, respectively. illustrated in Fig. 5.

*b)  1-DoF revolute plantarflexion control*

1-DoF revolute control represents a modified version of the gait-phase dependent impedance control scheme published in [24], which adjusts stiffness and equilibrium angle parameters to create net-positive mechanical work at the ankle joint during stance. The stance phase is divided into a loading and unloading phase, the transition between which is determined by a change in ankle joint velocity from increasing dorsiflexion to increasing plantarflexion. In the loading phase, the nonlinear virtual spring being simulated by the motor stores less potential energy than is released during the unloading phase. This results in ankle mechanics that replicate the function of the biological ankle during walking gait and provides positive mechanical power during push-off, typical of many revolute powered ankle-foot prostheses that primarily provide propulsion work [27], [28], [29], [30]. This finite state machine logic is implemented in Simulink and is illustrated in Fig 5.

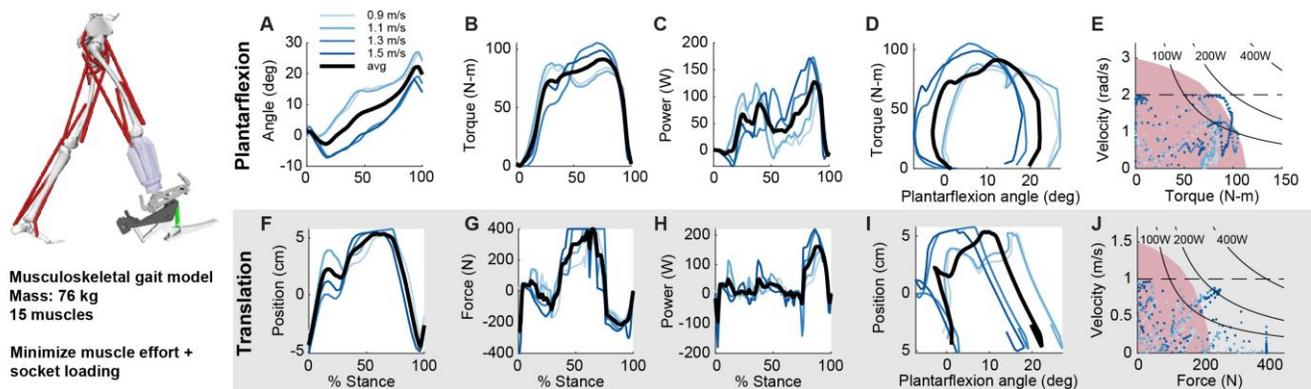

Figure 4. Optimal simulated prosthesis dynamics minimizing muscle effort and socket-residuum loading. Simulations were performed for 3 socket interface stiffness conditions x 4 walking speeds for a total of 12 conditions. Blue lines represent the average trajectory for each walking speed and the bold black line represents the overall average trajectory. Shaded regions in plot E and J represent the continuous operation region of the selected actuators and blue points represent simulated prosthesis dynamics at each time point of the simulation.

*c)  2-DoF simulation-based walking control*

For 2-DoF control, control targets attempt to replicate simulated prosthesis dynamics. The simulations are solutions to optimal control problems with 51 discrete control targets per stride, and while close tracking to simulated force-displacement curves is technically possible, a more generalizable approach may be to discern the functional role of each DoF from the mechanical work generated from each DoF for each phase of stance. Based on the power generation from the simulated revolute joint (Fig 4), its function can be explained as generating push-off power, consistent with the role of the biological ankle during walking [31], [32], [33]. Therefore, we simply continued to apply the 1-DoF revolute controller to the revolute actuator.

The role of the translation axis is more complicated. The simulated actuator outputs relatively low power on average during the majority of the stance phase, with a burst of positive power during push-off. The socket translates to the maximum anterior position during most of stance, maintaining a high static force to maintain this position, then reverses to translate backward relative to the foot to exert positive mechanical power. This is likely because anterior translation reduces the interaction moment between the socket and residual tibia, and an externally actuated posterior translation during push-off effectively increases the step length for no additional effort by the walker. Therefore, the translation axis is controlled to immediately translate forward and maintain this position during the loading phase of stance, then stiffly translate backward during the unloading phase, augmenting step length and propulsion (Fig 5).

*2) Position, Load, and Stiffness Control*

The stiffness controller is designed to make the prototype plantarflex with the dynamics of a torsion spring system with an adjustable spring constant. Output force and torque are commanded based on the following control law:

$$T = K(x - x_0) \quad ()$$

where $T$ is the commanded plantarflexion torque, K is the linear stiffness of the virtual torsion spring, $x$ is the measured plantarflexion angle, and $x_0$ is the virtual equilibrium plantarflexion angle. Inertial terms are not included, allowing the natural inertial properties of the device to govern its dynamic response.

The torque commanded by the impedance controller is output by the low-level torque controller. This controller is designed with a PD architecture and outputs a commanded motor velocity. This architecture has been shown to have smoother and higher bandwidth performance for series-elastic actuators than direct motor current or voltage control [34]. Closed loop velocity control is implemented internally by the Maxon ESCON 50/5 motor drivers. Torque control is linearized by calculating the cable tension corresponding with plantarflexion torque, given the current plantarflexion angle and geometric parameters of the cable connection to the prosthetic foot. Desired torque is converted into desired cable tension, which is used as the controller input, and which has a linear relationship with motor velocity.

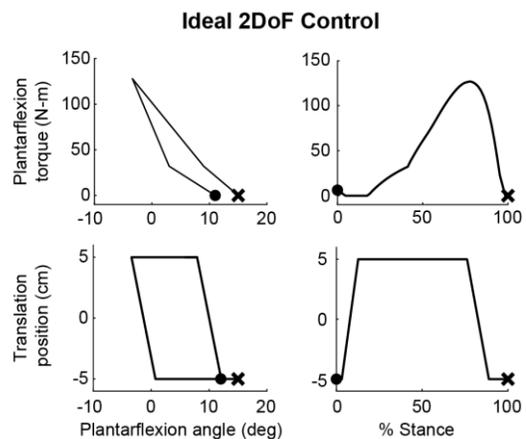

Figure 5. Ideal walking control behavior for both DoFs based on simulated optimum and existing prosthesis controllers. The left column depicts the control trajectory based on the state of the revolute ankle joint position. The right column depicts the control trajectory defined by this relationship over the duration of a stance phase for typical walking ankle kinematics. Filled circles mark heel strike events and crosses mark toe-off events.

Low-level position control is also accomplished using a PD controller architecture, and is used to control plantarflexion angle during swing and AP translation throughout the gait cycle. This controller is also used for both actuators when the prosthesis needs to rigidly hold a set position, such as during benchtop testing.

### III. EVALUATION

We evaluated the prosthesis by characterizing the load estimation accuracy, actuation limits, closed loop force/torque/position control bandwidth, and reference tracking during walking.

#### A. Benchtop Evaluation

Torque measurement accuracy was characterized as root mean squared (RMS) error between applied and measured torques by applying known torques about the flexion axis and known forces along the translation axis using free-hanging weights, as illustrated in Fig 6. Weights were sequentially added and removed to record the presence of any loading-unloading hysteresis. For this test, a fixed static position was controlled for each motor, allowing the end effector to deflect under load as the transmission deflected. Measured torque RMS error was 6.8 Nm (Fig. 6a) and force RMS error was 49.5 N (Fig 6b).

The following actuator characterization tests were performed by rigidly connecting the prosthesis end effector to a test fixture in series with a single-axis load cell (Futek, LCM325) to record output force. The test fixture couples the prosthesis to a set of extension springs which act as a compliant load (Fig. 7a). The fixture can be locked to create isometric loading conditions. The prosthesis-mounted encoders were used to record the output position and velocity. The load cell reading was converted to plantarflexion torque by computing the dot product of the moment arm and line of tension to the test fixture, using the plantarflexion angle data recorded by the prosthesis-mounted encoder and measurements of the experimental setup. The system was powered by a laboratory power supply providing 40V – about 83% of the nominal voltage rating of the motors. Each DoF was tested individually. The DoF not in use was controlled to maintain a fixed end-effector position for each test. Results for each test represent the average of 10 trials.

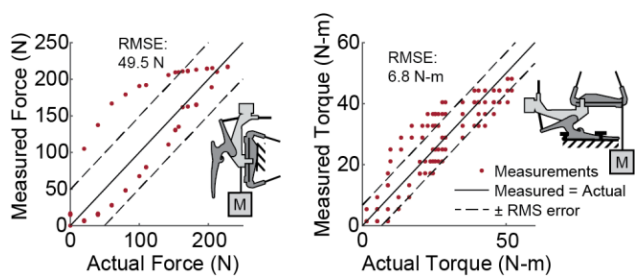

Figure 6. Loading measurement accuracy for (A) AP translation force and (B) plantarflexion torque.

Open loop actuator performance was characterized by measuring the peak output load, speed, and power for each DoF. These measures were quantified by attaching the prosthesis to the compliant load test configuration and generating a step input beyond the maximum capability of the actuator (Figure 7a). This test causes the prosthesis to smoothly traverse the performance range from no-load actuation to actuator stall. Output load and velocity were recorded for the duration of the test and multiplied to compute output power. Peak plantarflexion torque was measured at 160 N-m, peak angular velocity at 1.12 rad/s, and peak power at 122 W. Peak translation force was measured at 394 N, peak translation velocity at 0.4 m/s, and peak power at 140 W.

To perform closed-loop translation position and plantarflexion torque step response tests, the prosthesis was attached to the apparatus used in the previous test, which was locked in place (Fig. 7b). Plantarflexion torque was commanded as a square wave with a 5 second period, alternating between 0 and 100 N-m. AP translation was commanded as a 5 second square wave alternating between -4.5cm and +4.5 cm. We measured the 90% rise and fall time for each DoF. The rise and fall times for plantarflexion torque were 191 ms and 179 ms, respectively (Fig. 7b). Rise time for anterior and posterior translation position control was 162 ms and 159 ms, respectively (Fig. 7c).

We tested the closed-loop torque and position control bandwidth with the device fixed in the same configuration as the step response tests. Plantarflexion torque and AP translation were commanded as a logarithmic chirp signal from 0.1 to 30 Hz over 20 seconds. Plantarflexion torque oscillated between 20 and 70 N-m, and translation position oscillated between -3 and 3 cm. The measured and commanded values were converted into the frequency domain using a fast Fourier transform, and the magnitude ratio and phase difference across the frequency range were used to create a Bode plot. The bandwidth was determined using -3dB as the magnitude cutoff and 45° as the phase margin cutoff. Plantarflexion torque control bandwidth was 7.2 Hz and translation position control was 6.9 Hz, both limited by the -3dB crossover criterion (Figure 7D).

B. *Walking Evaluation*

We tested the performance of the device during walking to evaluate the controller performance during normal operating conditions. One test user without amputation (75.1 kg, 1.91 m tall, 32 years, male) wore the prosthesis using a set of adapter boots designed to immobilize the anatomical ankle and safely transmit the device loads through to the body. The test user walked with the prosthesis in revolute and simulation-optimized control modes for over 5 minutes at 1.0 m/s per controller. 50 strides with successful detection of gait events were selected to calculate the mean and variance of the walking controller trajectories. Kinematics and loads were measured for each degree of freedom during walking and compared with the target load/position relationships (Fig 8). RMS torque and position tracking errors are reported for each condition in Fig 8.

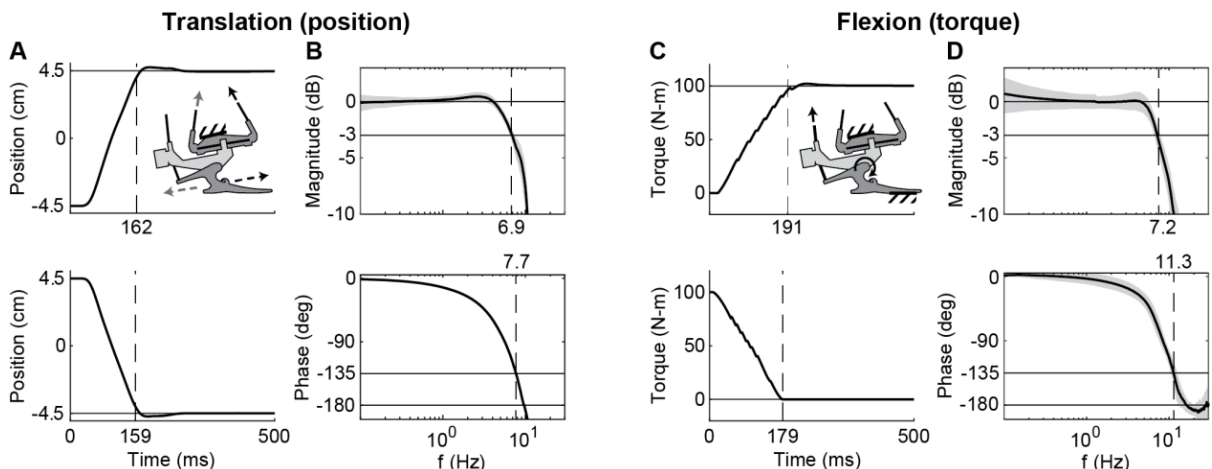

Figure 7. Actuator performance results: Bidirectional AP translation axis closed loop position control (A) step response and (B) frequency response, and closed loop plantarflexion torque control (C) step response and (D) frequency response.

## IV. Discussion

We developed a highly configurable 2 DoF ankle-foot prosthesis, capable of independently controlling plantarflexion and AP alignment. The prototype uses a modular, "wearable-offboard" actuation approach to transmit power to a distal joint while proximally concentrating the mass on the body and still enable untethered operation. We evaluated the device and found that its actuation power and bandwidth, while not as high as fully off-board emulator devices [24], [26], [35], are in-line with other mobile prostheses and capable of performing a range of functional control modes during walking.

### A. Series-elastic load estimation

As expected, estimating loads through the actuated degrees-of-freedom using series elasticity and endpoint deflection measurement is more effective when the series elasticity is explicitly designed to be measurable and linear, as is the case for loading in plantarflexion. 6.8 N-m of RMS plantarflexion moment measurement error is larger than would be expected with a traditional strain-gauge based measurement approach, but is a reasonable estimate for the loading expected during walking. This may no longer be the case, however, for scenarios requiring precise control of smaller torques, such as standing balance under perturbation. This error primarily comes from hysteresis caused by stiction between the cable and sheath, which is affected by the degree of bend in the cable and the actuation speed. Bend angle will also slightly affect the effective length of the cable as it configures into the shortest path allowed within the sheath under tension, which is a source of error not reflected in the static test.

The above sources of error are exacerbated for the estimation of translation force, which has a much stiffer transmission. Endpoint deflections are therefore smaller and provide a smaller signal-to-error ratio. As AP translation involves bidirectional actuation, unlike plantarflexion, cable tension must be maintained in both directions simultaneously or a dead-zone develops where the end-effector can deflect under nearly no load. In practice, tension in this axis needed to be repeatedly adjusted during testing, making estimation of the load unreliable.

Overall, series-elastic deformation can provide a reasonable estimate of load through the actuators by tuning the stiffness of the elastic component to provide appropriate resolution for the loads being measured, given known tradeoffs to control bandwidth [36], [37]. Depending on the accuracy of measurement required for control and for analysis of gait, strain-gauge measurement of load may be necessary, whether through commercial load cells or strategic application of strain gauges to the device structure itself.

### B. Actuation and control performance

As demonstrated by the gait experiments, the prototype meets the actuation requirements for enabling walking gait. The primary driver of error in the walking controllers is phase lag. For plantarflexion torque, this error does not seem to be negatively affected by the presence of active translation changing the effective moment arm – in fact, tracking performance slightly improved during 2DoF control. Most of the tracking error occurs during early stance, when commanded torque is lowest. If the first 15% of the stance phase is excluded, plantarflexion torque tracking RMS error falls to 4.3 and 2.8 N-m for revolute and 2DoF walking control, respectively.

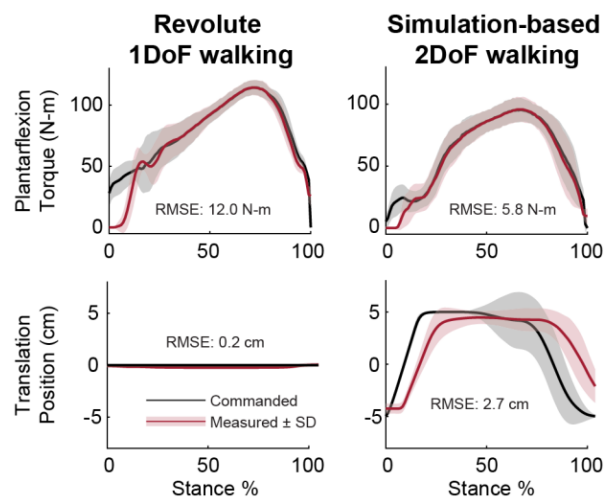

Figure 8. Performance of the closed loop controllers during walking for both the conventional revolute controller and the simulation-based 2DoF controller.

The translation position error does not similarly benefit from this analysis, however. While the translation axis is capable of maintaining a neutral locked position during revolute control, it lags significantly behind the reference signal when subjected to walking loads during 2DoF control, and this lag is apparent during early and late stance. The lower tracking accuracy of this axis is somewhat expected – the time constant of linearly positioning the inertial system of the human body is necessarily slower than increasing an interaction torque, especially considering that this operation is bi-directional. While it can achieve large translations during gait, the translation axis is limited in performance by the force-generating capability of its actuator. It was observed that translation could be stalled during walking if the wearer intentionally landed hard on the prototype and used the extended moment arm afforded by the boot adapters to exert a strong off-axis moment during stance. These loads should not typically be experienced during walking without an adapter, but the sensitivity of the translation axis to load disturbance indicates a limitation in the design.

This limitation is in large part driven by sizing the actuation system to be wearable on the body. While mechanical power output comparable to fully offboard systems (e.g. over 1 kW, [24]) is not realistic for such a system to be practical, there is likely further room for improvement. Small, high torque motors (e.g. AK80 series, T-Motor) are now available which reduce the need for multiple stages of speed reduction in the transmission, likely a major source of energy losses in this prototype. Furthermore, we acknowledge that a similar wearable cable actuated system has been developed (Portable Caplex System, Human Motion Technologies LLC) during the development of this report, though no technical specifications have been published for comparison.

*C. Sim to Real: Practical lessons and further opportunities*

A major objective of the prototype was to replicate prosthesis behaviors generated from optimal control biomechanical simulations, and our testing validated its ability to replicate one such set of simulations. Despite our focus in this paper on the performance of the prototype and not the biomechanics of the full human-robot system, these tests have revealed implications for biomechanical simulations as a design tool for wearable robots.

Optimal control simulations provide a useful reference for what may be biomechanically possible for users to achieve with a given device. However, using them to design actuation targets may lead to underestimates when simulations involve effort optimization. Simulated human-device interaction using optimal control can be considered a best-case scenario, in which the user and device operate in perfect harmony to minimize effort and interaction loads. As demonstrated, even for a user familiar with the theoretically optimal gait, interaction forces and actuator demands may exceed simulated behavior. Similarly, using biomechanical data from normal walking such as ground reaction force magnitudes or joint dynamics may not communicate the functional demands of an artificial, non-anatomical joint as the user learns to use it, potentially experiencing or even deliberately exploring higher load cases in the course of adaptation.

Particular care must be taken with degrees-of-freedom which can be easily backdriven by the wearer. Conservative factors of safety may need to be applied to the output load requirement for these actuators, or the designer risks a control axis which cannot maintain its position under load. In this prototype, the translation axis was designed with high position control bandwidth as the primary goal, and typical walking ground reaction forces in the AP direction are much smaller than their vertical components. However, if the user tests the range of motion of the prosthesis they can align the translation axis such that a significant portion of the vertical ground reaction load travels along it, which could cause this actuator to track poorly. This consideration applies particularly to prostheses – in many cases, exoskeleton actuators which can be overpowered by their users may be a desired feature.

The development of this sagittal planar prosthesis enables multiple avenues for future work. Toward improving prosthesis design, the gait biomechanics and energetics of participants with lower limb amputation wearing the device in response to different controllers can be analyzed to determine the ways in which configurable prosthesis dynamics influence the gait of those who use them. Furthermore, if the device behavior is configured to a particular measure, the influence of that behavior on that measure can be experimentally tested and modeled (e.g. metabolic energy expenditure, socket-residuum pressures). The independent control of push-off and alignment may also be utilized to study user preference between effort-saving and comfort-enhancing prosthesis behaviors.

Toward improving our understanding of motor learning and adaptation in prosthesis users, the change in the previously stated outcome measures, or other biomechanical markers, can be measured over time to assess the ability of participants to learn to exploit the device mechanics, if at all. Furthermore, we may investigate hypotheses on the driving phenomena behind this adaptation by comparing post-adaptation gait to gait predicted by various optimality criteria in simulation.

Finally, this platform enables the above studies to be conducted either on treadmills or overground to investigate differences in steady state walking versus more practical locomotion scenarios.

## V. Conclusion

We designed a prosthetic ankle-foot capable of recreating 2-DoF dynamics computed via optimal control gait simulations. This prototype is designed to function as an test platform for experimental investigation of the effect of novel prosthesis dynamics on relevant outcome measures such as metabolic energy expenditure or socket-residuum loading. The design compromises between the control bandwidth and scalability of offboard-actuated platforms and the mobility of untethered, self-contained platforms. In testing, the prosthesis demonstrated the ability to responsively and independently provide active plantarflexion and AP socket alignment during gait, as well as reproduce the behavior of existing revolute powered prosthetic ankle-feet. The development of this prototype enables novel experiments with prosthesis users and provides practical insights to using optimal control musculoskeletal simulations to design wearable robots.

Acknowledgment

We would like to thank Hari Krishnan, Chloe Smith, Sean Flanagan, and Abigail Risse for their assistance with device fabrication, characterization, and data collection.